\documentclass[letterpaper]{article} % DO NOT CHANGE THIS
\usepackage{aaai2027}  % DO NOT CHANGE THIS
\usepackage[hyphens]{url}  % DO NOT CHANGE THIS
\usepackage{graphicx} % DO NOT CHANGE THIS
\usepackage{natbib}  % DO NOT CHANGE THIS AND DO NOT ADD ANY OPTIONS TO IT
\usepackage{caption} % DO NOT CHANGE THIS AND DO NOT ADD ANY OPTIONS TO IT
\usepackage{algorithm}
\usepackage{algorithmic}

\usepackage{booktabs}

\usepackage{amsmath}
\usepackage{amssymb}
\usepackage{multirow}
\usepackage{tabularx}
\usepackage[table]{xcolor}
\usepackage{enumitem}

\title{UniGD: A Unified Generative-Discriminative Framework for Industrial Retrieval}

\copyrighttext{* Both authors contributed equally to this research. \\ $^\dagger$Corresponding author.}

\author{
    Shujie Ji$^{*}$,
    Yawei Kong$^{*}$,
    Yilin Zhao,
    Li Wang$^{\dagger}$,
    Xialong Liu,
    Peng Jiang \\
    {\normalfont Kuaishou Technology, Beijing, China} \\
    {\normalfont\small \{jishujie03, kongyawei, zhaoyilin05, wangli18, zhaolei16, jiangpeng\}@kuaishou.com}
}
\affiliations{}

\begin{document}

\maketitle

\begin{abstract}
Generative retrieval (GR) is a promising paradigm for industrial search advertising, yet its deployment is constrained by strict relevance and latency requirements. Existing systems cascade GR with an independent relevance model, decoupling the generative likelihood objective from query–ad relevance discrimination, which compromises effectiveness and increases serving costs. We propose a Unified Generative-Discriminative framework (UniGD) that integrates retrieval and relevance scoring within a single model. To mitigate gradient interference in joint optimization, UniGD introduces Conflict-Aware Gradient Enhancement (CAGE) to adaptively coordinate the two objectives. UniGD further designs a Codebook-Anchored Representation Module (CAM) that anchors item representations to frozen hierarchical codebooks distilled from a multimodal pretrained model, thereby endowing them with rich and generalizable semantic priors. For heterogeneous short-video, product, and live-stream ads, UniGD proposes Heterogeneous Ad-material Modeling (HAM), which captures cross-type semantic commonality over a shared backbone while preserving type-specific modeling capacity. Online A/B tests on Kuaishou search advertising platform show that UniGD raises ad revenue by 5.78\%, reduces inference latency by 33.1\%, and improves discriminative relevance estimation. On NQ320K and MS300K, UniGD improves Recall@10 over the strongest reproduced GR baseline by 8.44\% and 3.19\%, respectively.
\end{abstract}


\section{Introduction}
\label{sec:introduction}
Search advertising systems must retrieve semantically relevant ads from hundreds of millions of candidates under stringent latency constraints. Conventional multi-stage retrieval pipelines tackle this challenge through progressive candidate filtering, but each stage is typically optimized independently, leading to inconsistent objectives and limited overall retrieval effectiveness.
Generative Retrieval (GR)~\cite{tay2022transformer,bevilacqua2022autoregressive} has recently attracted increasing attention as a promising paradigm for large-scale candidate retrieval. Through autoregressive decoding, GR casts candidate retrieval as the conditional generation of Semantic Identifiers (SIDs)~\cite{vqvae2017,rajput2023recommender,deng2025onerec}. This generative paradigm moves beyond conventional matching-based retrieval and enables unified modeling of query-candidate semantic interactions.

In real-world industrial search advertising deployments, GR is typically cascaded with a downstream relevance model to ensure the relevance of the final results. Specifically, GR first retrieves top-$K$ candidate ads by autoregressive generation likelihood, and the relevance model then assigns discriminative relevance scores using query-ad matching signals. Despite its practical adoption, this cascaded paradigm still faces several key challenges in industrial deployment.

(1) \textbf{Objective discrepancy and serving overhead.}
The discrepancy between generative retrieval and discriminative relevance scoring may cause relevant ads with low generation probabilities to be prematurely discarded during top-$K$ truncation, before reaching the downstream relevance model for rescoring~\cite{tang2024gr2,cheng2025ddid}. Moreover, the cascaded GR paradigm introduces additional online inference latency. Although previous studies introduce ranking supervision to reshape the relative ordering of generation likelihoods~\cite{zhou2024roger,li2024ltrgr}, such preference signals do not directly produce explicit point-wise relevance scores. Consequently, they are unsuitable for threshold-based filtering and offer limited reliability as relevance features for downstream modules.

(2) \textbf{Representations for long-tail and new ads.}
Reliable relevance scoring requires discriminative ad-side representations, but long-tail and new ads often lack sufficient interaction signals for representation learning. This issue is particularly pronounced on our platform, where new ads account for as much as $36.9\%$ of daily traffic. Existing cascaded or auxiliary discriminative modules~\cite{tang2024gr2,onepiece2024} typically rely on separately trained ad encoders or periodically updated ad embeddings, making it difficult to provide reliable representations immediately after indexing.

(3) \textbf{Unified modeling of heterogeneous ad materials.}
Industrial search advertising covers multiple ad material types, including short-video, product, and live-stream ads, which exhibit significant semantic heterogeneity.  Existing GR methods lack explicit modeling of cross-material differences, and thus struggle to accommodate the semantic characteristics of different material types within a shared SID semantic space, limiting the generalization ability of GR in industrial advertising deployments.

To address these challenges, we propose \textbf{UniGD}, a \textbf{Uni}fied \textbf{G}enerative-\textbf{D}iscriminative framework that integrates generative retrieval and relevance discrimination within a single model. 
Unlike cascaded GR, UniGD jointly models autoregressive generation and query-ad relevance discrimination over a shared backbone, achieving mutual reinforcement and optimization synergy while eliminating the serving overhead of an additional relevance model. Specifically, UniGD introduces three key designs.
(1) \textbf{Conflict-Aware Gradient Enhancement (CAGE).}
Since jointly optimizing generative and discriminative objectives over a shared backbone may produce inconsistent gradient update directions, we introduce CAGE to coordinate their gradients during backpropagation, enabling the two objectives to benefit from joint optimization.
(2) \textbf{Codebook-Anchored Representation Module (CAM).}
To address the cold-start bottleneck in ad retrieval under sparse behavioral supervision, we design CAM, which anchors item-side representations to frozen hierarchical codebooks derived from a large-scale multimodal pretrained model, allowing long-tail and new ads to inherit its semantic priors and obtain reliable representations immediately upon indexing.
(3) \textbf{Heterogeneous Ad-material Modeling (HAM).}
Considering the semantic heterogeneity across ad materials, we propose HAM to assign each material type to a dedicated semantic space and inject material-aware signals into the unified backbone. In this way, both SID generation and relevance discrimination can focus on the appropriate signal sources for each material type, thereby enhancing the unified generative-discriminative process while avoiding an over-simplified homogeneous formulation.

We validate UniGD through both industrial-scale online deployment and public benchmark evaluation. Online A/B tests on Kuaishou search advertising platform demonstrate that UniGD improves ad revenue by $5.78\%$ while reducing inference latency by $33.1\%$, from 13.00 ms to 8.70 ms. On the NQ320K and MS300K benchmarks, UniGD outperforms the strongest reproduced GR baseline in Recall@10 by 8.44\% and 3.19\%, respectively.
The main contributions of this paper are summarized as follows:
\begin{itemize}[topsep=0pt,itemsep=0pt,parsep=0pt,partopsep=0pt,leftmargin=*]
\item We propose UniGD, a unified generative-discriminative retrieval framework that jointly performs autoregressive SID generation and explicit discriminative relevance scoring within a single shared
model, eliminating the need for a cascaded relevance model.

\item We develop three key components for effective industrial deployment: CAGE mitigates optimization conflicts between the generative and discriminative objectives; CAM provides reliable item-side representations for long-tail and newly indexed ads; and HAM captures semantic heterogeneity across short-video, product, and live-stream ads.

\item We conduct extensive online and offline evaluations, demonstrating that UniGD substantially improves retrieval effectiveness and advertising revenue while reducing online inference latency.
\end{itemize}

\section{Related Work}
\label{sec:related}
\subsection{Generative Methods}
Unlike traditional term-based~\cite{robertson1995bm25,nogueira2019doc2query,raffel2020t5} and dense retrievers~\cite{karpukhin2020dense,xiong2021ance,zhan2020repbert,ni2021sentence,rocketqa}, generative retrieval recasts large-scale candidate retrieval as sequence generation. Early studies established the technical foundation via document-identifier generation, query-generation augmentation, prefix-aware decoding, and learnable tokenization~\cite{tay2022transformer,zhuang2022dsiqg,wang2022neural,sun2023gentoken}. Subsequent work advanced the paradigm through scalable multi-stage training, efficient simultaneous decoding, and semantic docid design~\cite{zeng2024ripor,zeng2024pag,zhou2022ultron,li2024genrpo,onesug}, and extended SID-based generation to sequential recommendation and industrial recommendation and advertising pipelines~\cite{rajput2023recommender,deng2025onerec,zheng2025egav2,gr4ad2024}. Despite this progress, existing GR methods largely optimize generation probability via MLE and rely on beam search at inference, which inflates high-frequency SIDs and underestimates semantically relevant long-tail and new items.

\subsection{Generative-Discriminative Methods}
Several studies integrate ranking-oriented or discriminative signals into generative retrieval and recommendation~\cite{unigrf,grank,rankgr,chen2025onesearch}. Representative efforts introduce ranking-oriented objectives or document-level relevance optimization into MLE-based GR~\cite{zhou2024roger,li2024ltrgr,mekonnen2025ddro}, align generation with query-item relevance through end-to-end training~\cite{pang2025gram}, and design discriminative identifiers to sharpen candidate distinction~\cite{cheng2025ddid}. While these works make valuable progress in bridging generation with relevance-oriented ranking, search advertising's rapid ad turnover demands higher retrieval and discrimination accuracy, calling for a unified framework that tightly couples candidate generation with relevance estimation while ensuring scalability.

\section{Methodology}
\label{sec:method}
\subsection{Overview}
\label{sec:overview}
\begin{figure*}[t]
  \centering
  \includegraphics[width=\textwidth]{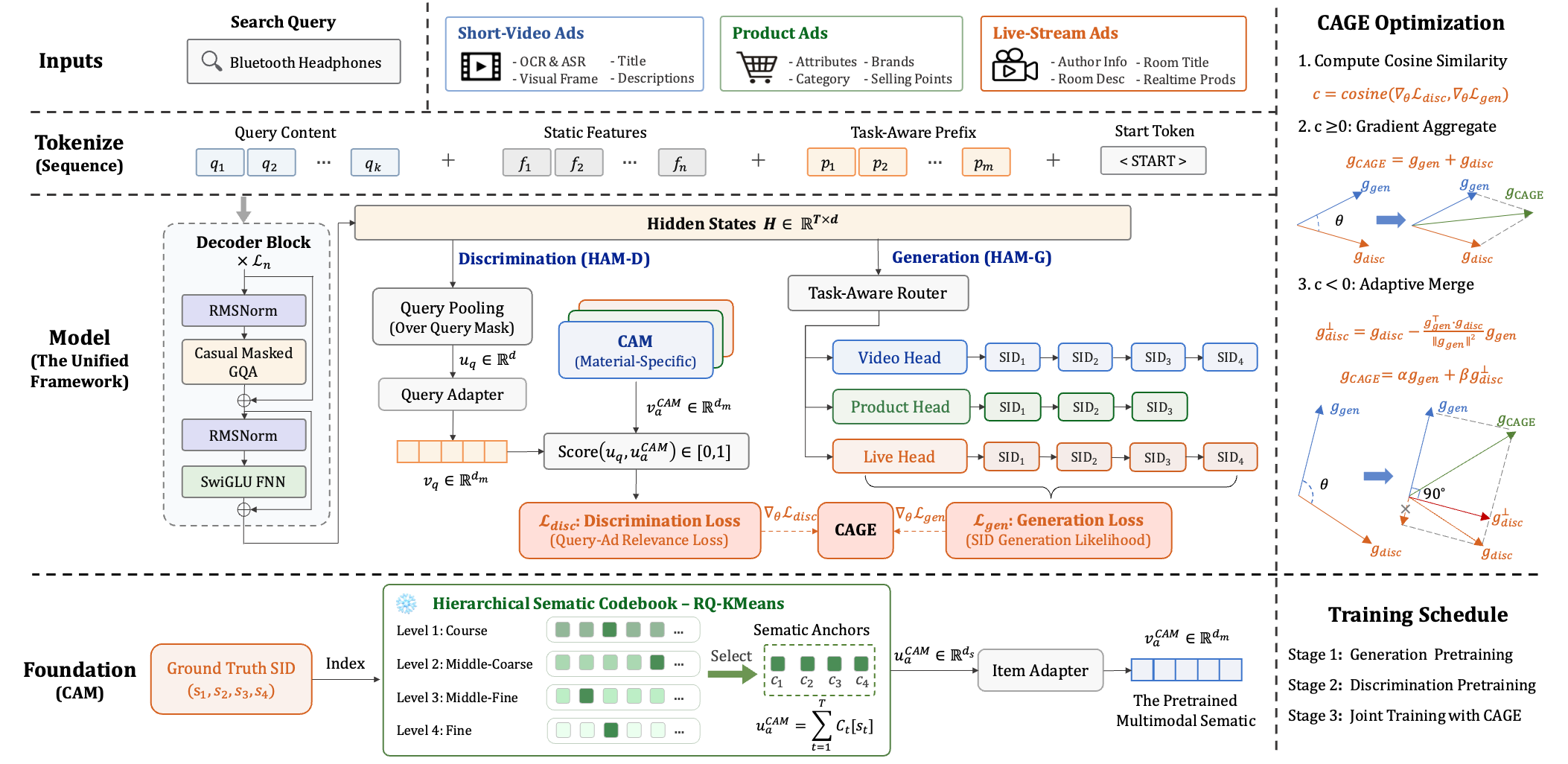}
  \caption{Training framework of UniGD. HAM-G and HAM-D jointly optimize the generative and CAM-based discrimination, while CAGE coordinates their gradients through a three-stage training scheme.}
\label{fig:model_arch}
\end{figure*}

In this paper, we propose UniGD, a unified decoder-only framework that integrates SID-based candidate generation and query-ad relevance scoring. As illustrated in Figure~\ref{fig:model_arch}, HAM captures material-specific semantics for both generative and discriminative learning, CAM derives codebook-anchored ad representations from frozen SID codebooks for relevance scoring, and CAGE coordinates the two objectives during joint optimization. Figure~\ref{fig:system_position} presents the corresponding industrial serving architecture, where candidate generation and relevance scoring are performed within the same model, eliminating the external online relevance model and thereby reducing inference latency and serving overhead.

\subsection{Unified Generative-Discriminative Modeling}
\label{sec:unified_modeling}
Given a query $q$ and an ad corpus $\mathcal{C}$, UniGD unifies
autoregressive SID generation and discriminative relevance estimation
within a shared decoder-only backbone. It jointly models the
conditional generation probability $p(\mathbf{s}(a) | q)$ and
the explicit relevance score $r(q,a)$.

\emph{SID tokenization.}
Each ad $a\in\mathcal{C}$ is indexed by a hierarchical semantic
identifier (SID)
$\mathbf{s}(a)=(s_1(a),\ldots,s_T(a))$ constructed using RQ-KMeans.
For autoregressive generation, we augment the model vocabulary with
level-specific SID tokens and serialize $\mathbf{s}(a)$ as a token
sequence, where $s_t(a)\in\mathcal{V}_t$ denotes the SID token at
level $t$. 
The same SID indices are used by CAM to
retrieve the corresponding codewords for relevance
estimation.

\emph{Autoregressive SID generation.}
Conditioned on a query $q$, UniGD factorizes the SID generation
probability as
\begin{equation}
p(\mathbf{s}(a)\mid q)
=
\prod_{t=1}^{T}
p\left(
s_t(a)\mid q,\mathbf{s}_{<t}(a)
\right),
\label{eq:sid_factorization}
\end{equation}
where $\mathbf{s}_{<t}(a)$ denotes the preceding SID prefix. At
decoding step $t$, the shared decoder produces the last-layer hidden
state $\mathbf{h}^{\mathrm{sid}}_t\in\mathbb{R}^{d}$ and predicts the
next SID token over $\mathcal{V}_t$ as
\begin{equation}
\scalebox{1.0}{$\displaystyle
p\!\left(s_t(a)\mid q,\mathbf{s}_{<t}(a)\right)
=
\operatorname{Softmax}\!\left(
\mathbf{W}_t\mathbf{h}^{\mathrm{sid}}_t+\mathbf{b}_t
\right)_{s_t(a)}
$}.
\label{eq:sid_generation}
\end{equation}
where
$\mathbf{W}_t\in\mathbb{R}^{|\mathcal{V}_t|\times d}$ and
$\mathbf{b}_t\in\mathbb{R}^{|\mathcal{V}_t|}$
are the prediction parameters associated with SID level $t$, which are further adapted by HAM to capture
material-specific semantics.
The generative objective for a training pair $(q,a)$ is
\begin{equation}
\mathcal{L}_{\mathrm{gen}}
=
-\sum_{t=1}^{T}
\log p\left(
s_t(a)\mid q,\mathbf{s}_{<t}(a)
\right).
\label{eq:gen_loss}
\end{equation}

\emph{Discriminative relevance estimation.}
UniGD derives the query representation for relevance estimation from
the query-token hidden states of the shared decoder. Let
$\mathbf{H}_q\in\mathbb{R}^{N_q\times d}$
denote the last-layer hidden-state matrix at the $N_q$ valid
query-token positions. We aggregate these states into a query-level representation
$\mathbf{u}_q=\operatorname{Agg}(\mathbf{H}_q)$, where
$\operatorname{Agg}(\cdot)$ is instantiated as mean pooling in our
implementation. For a candidate $a$, let
$\mathbf{u}^{\mathrm{CAM}}_a\in\mathbb{R}^{d_s}$
denote its codebook-anchored representation constructed by CAM.
To align $\mathbf{u}_q$ and $\mathbf{u}^{\mathrm{CAM}}_a$ in a shared
semantic space, we employ a query-side adapter $g_q$ and an ad-side adapter
$g_a$, respectively. The relevance logit is then computed as
\begin{equation}
r(q,a)
=
\psi\!\left(
\left[
g_q(\mathbf{u}_q);
g_a(\mathbf{u}_a^{\mathrm{CAM}})
\right]
\right),
\label{eq:relevance_score}
\end{equation}
where $[\cdot\,;\cdot]$ denotes vector concatenation and $\psi$ is a
lightweight scoring network that outputs $r(q,a)\in[0,1]$.

Given a supervised candidate set
$\mathcal{D}_q=\{(a_i,y_i)\}_{i=1}^{K}$ for query $q$, where $y_i\in[0,1]$ is the normalized relevance target derived from
human annotation or an offline relevance model,
the pointwise loss is
\begin{equation}
\mathcal{L}_{\mathrm{point}}
=
\frac{1}{K}
\sum_{i=1}^{K}
\left(
r(q,a_i)-y_i
\right)^2.
\label{eq:pointwise_loss}
\end{equation}
To further enforce relative relevance ordering, we introduce an
auxiliary pairwise ranking loss. Given a positive ad $a^{+}$ and $M$
negative ads $\{a^{-}_j\}_{j=1}^{M}$ for query $q$, it is defined as
\begin{equation}
\mathcal{L}_{\mathrm{pair}}
=
\frac{1}{M}
\sum_{j=1}^{M}
\max\left(
0,
\gamma-r(q,a^{+})+r(q,a^{-}_j)
\right),
\label{eq:pairwise_loss}
\end{equation}
where $\gamma\in(0,1)$ is the ranking margin.
The complete discriminative objective combines pointwise score
supervision and pairwise ranking:
\begin{equation}
\mathcal{L}_{\mathrm{disc}}
=
\mathcal{L}_{\mathrm{point}}
+
\lambda\mathcal{L}_{\mathrm{pair}},
\label{eq:disc_loss}
\end{equation}
where $\lambda$ controls the contribution of pairwise supervision.

To fully develop both generative and discriminative capabilities,
UniGD adopts a three-stage training scheme. Stage$1$ and Stage$2$ separately optimize $\mathcal{L}_{\mathrm{gen}}$ and $\mathcal{L}_{\mathrm{disc}}$ to
initialize the corresponding task components. Stage$3$ jointly
optimizes both objectives with CAGE coordinating their gradients over
the shared parameters.

\subsection{Codebook-Anchored Representation Module}
\label{sec:cam}

Discriminative relevance estimation requires informative ad-side
representations under strict online latency constraints, which is particularly
challenging for new and long-tail ads with sparse interaction signals. To
address this issue, we introduce the \emph{Codebook-Anchored Representation
Module (CAM)}. Given an ad's SID, CAM constructs a quantized approximation
of its multimodal representation by looking up and combining the
corresponding codewords.

\emph{Multimodal residual quantization.}
We employ a large-scale multimodal encoder, pretrained on a large corpus of
ad creatives, to encode the heterogeneous content of each ad into an
informative semantic representation:
\begin{equation}
\mathbf{v}_a
=
\operatorname{Enc}_{\mathrm{MM}}(a)
\in\mathbb{R}^{d_s}.
\label{eq:multimodal_representation}
\end{equation}
RQ-KMeans is trained offline on the ad representations in the indexing
corpus to learn $T$ residual codebooks. Let
$\mathbf{C}_t\in\mathbb{R}^{|\mathcal{V}_t|\times d_s}$ denote the
codebook at level $t$. Given the residual
$\boldsymbol{\rho}_{t-1}(a)$ left by the preceding levels, the SID
token at level $t$ is assigned to its nearest codeword:
\begin{equation}
s_t(a)
=
\arg\min_{k\in\mathcal{V}_t}
\left\|
\boldsymbol{\rho}_{t-1}(a)-\mathbf{C}_t[k]
\right\|_2^2,
\label{eq:residual_quantization}
\end{equation}
where $\boldsymbol{\rho}_0(a)=\mathbf{v}_a$. Earlier levels capture
the dominant components of the multimodal representation, while
subsequent levels quantize the remaining residuals to progressively
encode finer-grained information.

\emph{Codebook-anchored reconstruction.}
CAM retrieves the codeword indexed by the SID token at each level,
$\mathbf{c}_t(a)=\mathbf{C}_t[s_t(a)]$, and recursively updates the
residual as
$\boldsymbol{\rho}_t(a)=
\boldsymbol{\rho}_{t-1}(a)-\mathbf{c}_t(a)$.
The ad-side representation is reconstructed by summing the selected
codewords:
\begin{equation}
\mathbf{u}^{\mathrm{CAM}}_a
=
\sum_{t=1}^{T}\mathbf{c}_t(a).
\label{eq:cam_representation}
\end{equation}
Consequently, $\mathbf{v}_a= \mathbf{u}^{\mathrm{CAM}}_a+\boldsymbol{\rho}_T(a)$, where $\boldsymbol{\rho}_T(a)$ denotes the final quantization residual. Thus, $\mathbf{u}^{\mathrm{CAM}}_a$ provides a quantized reconstruction of the original multimodal representation and serves as the ad-side input for discriminative relevance estimation. The codebooks remain frozen throughout UniGD training, preserving the correspondence between SID tokens and their offline-learned codewords and thereby providing stable semantic anchors. Once an ad is assigned an SID, CAM requires only $T$ codebook lookups and a lightweight vector summation, enabling relevance estimation immediately after indexing without online ad-side encoder inference or complex feature construction.

\subsection{Heterogeneous Ad-material Modeling}
\label{sec:ham}

Industrial search advertising contains heterogeneous ad materials,
including short-video, product, and live-stream ads. They differ
substantially in content modalities, structures, presentation formats,
and information density, resulting in different semantic distributions
and relevance patterns. A homogeneous modeling scheme may therefore
obscure material-specific semantics. We introduce \emph{Heterogeneous Ad-material Modeling (HAM)} to specialize codebook construction, SID
generation, and relevance estimation by material type while retaining the shared backbone.

Let $m\in\mathcal{M}=\{\mathrm{video},\mathrm{product},\mathrm{live}\}$
denote the material type. HAM performs residual quantization separately
for each type, yielding material-specific codebooks
$\{\mathbf{C}_t^{(m)}\}_{t=1}^{T_m}$ and SID vocabularies
$\{\mathcal{V}_t^{(m)}\}_{t=1}^{T_m}$. For generation, it replaces the
output parameters in Eq.~\ref{eq:sid_generation} with
$\mathbf{W}_t^{(m)}$ and $\mathbf{b}_t^{(m)}$, which predict over
$\mathcal{V}_t^{(m)}$. During training, $m$ is obtained from the target
ad's material-type metadata. At inference, all material-specific heads
perform type-specific beam search in parallel using the shared decoder
hidden states, and the resulting typed SID candidates are merged for
relevance scoring without additional backbone forward passes. For
relevance estimation, CAM retrieves codewords from the codebooks of type
$m$ and reconstructs the material-aware ad representation
\begin{equation}
\mathbf{u}_a^{\mathrm{CAM},(m)}
=
\sum_{t=1}^{T_m}
\mathbf{C}_t^{(m)}[s_t(a)].
\label{eq:ham_cam_representation}
\end{equation}
Substituting $\mathbf{u}_a^{\mathrm{CAM},(m)}$ in the relevance scoring function
yields the material-aware relevance score $r(q,a,m)$. In this way,
generation and discrimination operate on consistent material-specific SID
semantics while sharing the backbone and scoring architecture across
materials.

\subsection{Conflict-Aware Gradient Enhancement}
\label{sec:joint_opt}

Joint optimization may produce conflicting gradients from the
generative and discriminative objectives over the shared decoder
parameters, leading to optimization interference. We therefore
introduce \emph{Conflict-Aware Gradient Enhancement (CAGE)} to
adaptively coordinate their gradients while prioritizing
autoregressive SID generation.

\begin{figure*}[t]
\centering
\includegraphics[width=\textwidth]{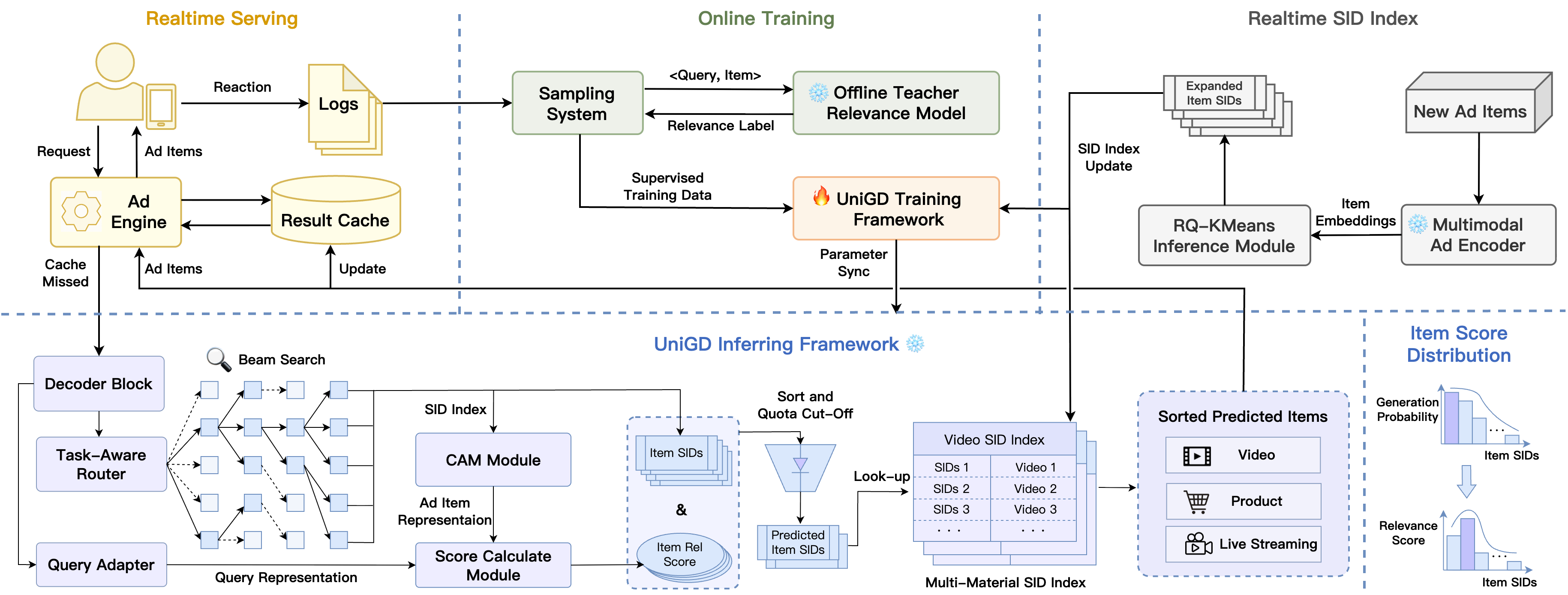}
\caption{Online serving pipeline of UniGD in industrial search advertising.}
\label{fig:system_position}
\end{figure*}

Let $\theta_s$ denote the shared decoder parameters. We compute the
task gradients $\mathbf{g}_{\mathrm{gen}}
=\nabla_{\theta_s}\mathcal{L}_{\mathrm{gen}}$
and
$\mathbf{g}_{\mathrm{disc}}
=\nabla_{\theta_s}(\eta\mathcal{L}_{\mathrm{disc}})$,
where $\eta>0$ rescales the discriminative gradient to reduce the
magnitude imbalance between the two objectives. Their directional
agreement is measured by the cosine similarity
$c=\operatorname{cos}(\mathbf{g}_{\mathrm{gen}},
\mathbf{g}_{\mathrm{disc}})$.
When $c\geq0$, the gradients are compatible and directly combined.
When $c<0$, CAGE removes the conflicting component of the
discriminative gradient by projecting it onto the orthogonal
complement of the generative direction:
\begin{equation}
\mathbf{g}_{\mathrm{disc}}^{\perp}
=
\mathbf{g}_{\mathrm{disc}}
-
\frac{
\mathbf{g}_{\mathrm{gen}}^{\top}\mathbf{g}_{\mathrm{disc}}
}{
\|\mathbf{g}_{\mathrm{gen}}\|_2^2
}
\mathbf{g}_{\mathrm{gen}}.
\label{eq:cage_projection}
\end{equation}
The gradient for updating the shared decoder is
\begin{equation}
\mathbf{g}_{\mathrm{CAGE}}
=
\begin{cases}
\mathbf{g}_{\mathrm{gen}}+\mathbf{g}_{\mathrm{disc}},
& c\geq0,\\[3pt]
\alpha(c)\mathbf{g}_{\mathrm{gen}}
+\beta(c)\mathbf{g}_{\mathrm{disc}}^{\perp},
& c<0,
\end{cases}
\label{eq:cage}
\end{equation}
where $\alpha(c)=1+\kappa(-c)$ and
$\beta(c)=1-\kappa(-c)$, with $\kappa\in[0,1]$ controlling the
adaptation strength. As $c$ decreases, CAGE increasingly favors the
generative direction, preserving SID generation while retaining
non-conflicting discriminative supervision. The resulting gradient is
used to update the shared decoder, while task-specific parameters are
optimized by their respective objectives.

\subsection{Industrial Online Serving}
\label{sec:serving}
Figure~\ref{fig:system_position} illustrates UniGD's continual learning, incremental indexing, and real-time serving pipeline deployed in our search advertising system. Request contexts and user feedback are collected to form query-ad training samples, with fine-grained relevance targets provided by an offline teacher model. The trained parameters are periodically updated on the inference servers. Meanwhile, the real-time SID index maintains material-specific SID-ad mappings. New or updated ads are quantized into SIDs and incrementally indexed. Since these mappings are independent of model updates, newly indexed ads can enter retrieval without rebuilding a model-dependent embedding index.

During serving, UniGD generates candidate SIDs via beam search and derives the query representation from the shared decoder. CAM reconstructs each candidate's ad-side representation from its material-specific codebooks, and the lightweight scoring module reranks the candidates using explicit relevance scores. The retained SIDs are then resolved to actual ads through the real-time index. This unified pipeline removes the need for both a separately deployed online relevance model and per-candidate ad-side encoder inference.

\begin{table*}[t]
\centering
\small
\setlength{\tabcolsep}{3.5pt}
\begin{tabular*}{\textwidth}{@{\extracolsep{\fill}}lcccccccc}
\toprule
\textbf{Model}
& \multicolumn{4}{c}{\textbf{NQ320K}}
& \multicolumn{4}{c}{\textbf{MS300K}} \\
\cmidrule(lr){2-5} \cmidrule(lr){6-9}
& \textbf{Recall@1} & \textbf{Recall@5} & \textbf{Recall@10} & \textbf{MRR@10}
& \textbf{Recall@1} & \textbf{Recall@5} & \textbf{Recall@10} & \textbf{MRR@10} \\
\midrule
DSI (SI)                              & 27.42 & 47.26 & 56.58 & 34.31 & 25.74 & 43.58 & 53.84 & 33.92 \\
NCI (SI)                              & 32.69 & 55.82 & 69.20 & 42.84 & 29.54 & 57.99 & 67.28 & 40.46 \\
Ultron (TU)                           & 33.78 & 54.20 & 67.05 & 42.51 & 29.82 & 60.39 & 68.31 & 42.53 \\
LTRGR (SI)                            & 32.80 & 56.20 & 68.74 & 44.80 & 32.69 & 64.37 & 72.43 & 47.85 \\
GenRRL (Sum)$^{*}$                   & 35.82 & 57.04 & \underline{71.49} & 45.93 & 32.78 & 63.91 & \underline{75.80} & 45.98 \\
DDRO (TU)$^{*}$                     & 40.52 & 52.81 & 55.63 & 45.68 & \underline{37.92} & \underline{66.13} & 73.72 & \underline{49.74} \\
DDRO (PQ)$^{*}$                     & \underline{48.57} & \underline{63.78} & 66.94 & \underline{55.18} & 32.61 & 64.05 & 72.71 & 45.43 \\
\midrule
UniGD (Ours)                                 & \textbf{72.33}$^{\dagger}$ & \textbf{75.84}$^{\dagger}$ & \textbf{77.52}$^{\dagger}$ & \textbf{74.21}$^{\dagger}$ & \textbf{60.64}$^{\dagger}$ & \textbf{71.85}$^{\dagger}$ & \textbf{78.22}$^{\dagger}$ & \textbf{65.12}$^{\dagger}$ \\
\bottomrule
\end{tabular*}
\caption{Generative retrieval performance (\%) on NQ320K and MS300K with representative generative retrieval baselines. \textbf{Bold}: best; \underline{underlined}: second best. $*$ denotes results reproduced under our unified protocol. $\dagger$ indicates a statistically significant improvement over the strongest reproduced baseline for each metric under a two-tailed paired $t$-test over per-query metric values ($p<0.05$). SI, TU, and PQ denote Semantic ID, Title and URL, and Product Quantization, respectively.}
\label{tab:results_public}
\end{table*}

\section{Experiments}
\label{sec:experiments}
\subsection{Experimental Settings}
\emph{Datasets.}
We evaluate UniGD on two public generative retrieval benchmarks and
one industrial search advertising dataset. NQ320K, derived from
Natural Questions~\cite{kwiatkowski2019natural}, contains $320$K
training queries and $7{,}830$ test queries over Wikipedia articles.
MS300K~\cite{zhou2022ultron}, curated from MS
MARCO~\cite{bajaj2016msmarco}, contains $320$K documents, $360$K
training queries, and $772$ test queries. We follow the data splits
used by DDRO~\cite{mekonnen2025ddro} for comparison with prior
work. For public benchmarks, query-document pairs are labeled as relevant (positive) or non-relevant (negative).
Our industrial dataset covers product, short-video, and live-stream
ads collected from our production search advertising logs. It contains $153$M
positive pairs for generative training and $248$M pairs for
discriminative training. The test sets contain $277$K pairs over $1$K queries and over $10$K
human-labeled pairs, respectively.

\emph{Evaluation Metrics.}
For generative retrieval, we report Recall@1/5/10 and MRR@10,
following prior work~\cite{wang2022neural,zhou2023genrrl,
zhou2024roger,zhou2022ultron}. On public benchmarks, significance
is assessed against the strongest reproduced baseline using a
two-tailed paired $t$-test over per-query values
($p<0.05$).
For relevance estimation on our industrial dataset, we
report AUC, Spearman and Pearson to evaluate binary discrimination,
rank consistency and score correlation. Relevance labels range from $0$ (irrelevant) to $3$ (highly relevant);
for AUC, labels $0$-$1$ are treated as negative and $2$-$3$ as
positive.

\emph{Baselines.}
For the industrial evaluation, we compare UniGD with the production
cascaded system, which performs generative retrieval followed by an
independently deployed relevance model. On the public benchmarks,
we compare UniGD with representative generative retrieval methods. DSI~\cite{tay2022transformer} formulates
retrieval as document-identifier generation; NCI~\cite{wang2022neural}
uses semantically structured numeric identifiers; Ultron~\cite{zhou2022ultron} combines keyword and semantic document identifiers with multi-stage training; LTRGR~\cite{li2024ltrgr} incorporates ranking supervision into
generative retrieval, GenRRL~\cite{zhou2023genrrl} optimizes generation using relevance feedback, and DDRO~\cite{mekonnen2025ddro} combines PQ-based identifiers with direct preference optimization~\cite{rafailov2023dpo}.

\emph{Implementation Details.}
For the public benchmarks, we instantiate UniGD with T5-base
($220$M parameters) and adopt the PQ-based SIDs of
DDRO with $T=24$. We reuse the frozen PQ centroids from DDRO as CAM codebooks. For our industrial setting,
we use a six-layer decoder-only Transformer with approximately $60$M
parameters, while the discriminative components introduce only
$1.12$M additional trainable parameters.
Our industrial SIDs use three quantization levels for product ads and
four for short-video and live-stream ads. Across both settings, UniGD
is trained with AdamW for two epochs in each stage. Stage~I optimizes
$\mathcal{L}_{\mathrm{gen}}$ with a learning rate of
$1\times10^{-4}$ and a batch size of $256$. Stage~II freezes the
shared decoder and optimizes only the query-side and item-side adapters
with $\mathcal{L}_{\mathrm{disc}}$, using a learning rate of
$1\times10^{-3}$, $\lambda=0.5$, and $\gamma=0.2$. Stage~III jointly
optimizes both objectives using CAGE, with $\eta=100$ and $\kappa=0.5$,
where $\eta$ balances their gradient magnitudes.
The learning rates are $5\times10^{-5}$ for the shared decoder and
$1\times10^{-4}$ for the task-specific modules. Industrial training takes three days on eight A800 GPUs, with beam widths of $100$  and $32$ for public and industrial inference, respectively.

\subsection{Public Benchmark Evaluation}
\label{sec:public_evaluation}
Table~\ref{tab:results_public} compares UniGD with representative
generative retrieval methods on NQ320K and MS300K. UniGD consistently
achieves the best performance across metrics, with statistically
significant improvements over the strongest reproduced baselines.
UniGD improves Recall@1 by more than $22$ absolute points on
both datasets, while yielding relative Recall@10 gains of $8.44\%$ on
NQ320K and $3.19\%$ on MS300K. The improvements in
Recall@1 and MRR@10 indicate that UniGD more reliably places relevant
candidates at the top of the ranking, highlighting the benefit of
explicit discriminative supervision for learning fine-grained
relevance signals between queries and candidates. Meanwhile, the
consistent gains in Recall@10 suggest that the benefit extends beyond
top-rank refinement to improved candidate coverage through the shared
generative backbone. The results on both benchmarks demonstrate the
effectiveness and generalizability of unified generative
and discriminative modeling.

\begin{table}[t]
\centering
\setlength{\tabcolsep}{4pt}
\small
\begin{tabular}{lcccc}
\toprule
\textbf{Variant}
& \textbf{Recall@1}
& \textbf{Recall@5}
& \textbf{Recall@10}
& \textbf{MRR@10} \\
\midrule

\multicolumn{5}{l}{\textit{Our industrial test set}} \\
UniGD
    & \textbf{19.57}
    & \textbf{32.83}
    & \textbf{53.37}
    & \textbf{27.20} \\
w/o CAM
    & 13.24
    & 26.67
    & 46.33
    & 20.21 \\
w/o HAM
    & 18.86
    & 32.13
    & 52.43
    & 26.39 \\
w/o CAGE
    & 18.36
    & 31.23
    & 51.57
    & 25.62 \\
Baseline
    & 12.53
    & 25.82
    & 43.98
    & 19.14 \\
    
\midrule
\multicolumn{5}{l}{\textit{NQ320K}} \\
UniGD
    & \textbf{72.33}
    & \textbf{75.84}
    & \textbf{77.52}
    & \textbf{74.21} \\
w/o CAM
    & 45.52
    & 58.96
    & 65.26
    & 52.03 \\
w/o CAGE
    & 67.22
    & 71.45
    & 74.38
    & 69.18 \\
Baseline
    & 42.50
    & 56.63
    & 61.18
    & 48.62 \\
\bottomrule
\end{tabular}
\caption{Comparison and ablation results on our industrial test set
and NQ320K.}
\label{tab:ablation}
\end{table}

\begin{table*}[t]
\setlength{\tabcolsep}{4.5pt}
\renewcommand{\arraystretch}{1.2}
\centering
\small
\begin{tabular}{l|cccccccc}
\hline
\textbf{Model Setting}
& \textbf{$\Delta$ Revenue}
& \textbf{$\Delta$ Impressions}
& \textbf{$\Delta$ Imp.\ (LT)}
& \textbf{$\Delta$ Imp.\ (New)}
& \textbf{$\Delta$ CTR}
& \textbf{ CVR}
& \textbf{Irrelevant Ratio}
& \textbf{Latency} \\
\hline
Baseline
& --
& --
& --
& --
& --
& --
& 6.79\%
& 13.0\,ms \\
GR+RelToken
& --2.08\%
& +1.84\%
& --3.42\%
& --5.71\%
& --0.25\%
& --0.24\%
& 17.30\%
& 10.1\,ms \\
UniGD-Gen
& +3.54\%
& +2.61\%
& +10.40\%
& +15.30\%
& +0.56\%
& +0.41\%
& 6.15\%
& 12.9\,ms \\
\textbf{UniGD}
& \textbf{+5.78\%}
& \textbf{+3.12\%}
& \textbf{+16.80\%}
& \textbf{+24.60\%}
& \textbf{+0.94\%}
& \textbf{+0.68\%}
& \textbf{5.81\%}
& \textbf{8.7\,ms} \\
\hline
\end{tabular}
\caption{Online A/B results in the Kuaishou search advertising system.
$\Delta$ denotes the relative change from the production baseline.
\textbf{Imp.\ (LT)} and \textbf{Imp.\ (New)} denote impression changes for
long-tail and new ads.  }
\label{tab:online_ab}
\end{table*}

\begin{table}[t]
\centering
\small
\setlength{\tabcolsep}{7pt}
\begin{tabular}{lccc}
\toprule
\textbf{Model Setting}
    & \textbf{AUC}
    & \textbf{Spearman}
    & \textbf{Pearson} \\
\midrule
Offline Teacher
    & \textbf{0.907} & \textbf{0.805} & 0.810 \\
\midrule
Online Student
    & 0.881 & 0.783 & 0.784 \\
GR+RelToken
    & 0.890 & 0.795 & 0.796 \\
\textbf{UniGD}
    & 0.905 & 0.798 & \textbf{0.811} \\
\bottomrule
\end{tabular}
\caption{Relevance estimation performance on our industrial
human-annotated test set. The offline teacher is included as a
reference and is not deployed for online inference.}
\label{tab:offline_corr}
\end{table}

\subsection{Industrial Offline Evaluation}
\label{sec:industrial_offline}
We evaluate UniGD on our industrial dataset from two complementary
perspectives: generative retrieval and discriminative relevance
estimation.

\emph{Generative retrieval.}
Table~\ref{tab:ablation} compares UniGD with a GR baseline using the
same backbone and SID configuration but trained only for autoregressive
SID generation. On our industrial test set, UniGD improves Recall@1
and MRR@10 by $7.04$ and $8.06$ points, respectively, demonstrating
the benefit of jointly modeling candidate generation and explicit
discriminative relevance estimation. Through the shared decoder, the
discriminative objective complements token-level generation with
relevance supervision, encouraging query representations to capture
informative query–candidate matching signals.

\emph{Relevance estimation.}
We evaluate relevance prediction on our human-annotated industrial
test set against a Qwen2-VL-7B offline teacher, a six-layer Transformer
deployed as the online student, and \emph{GR + RelToken}, which
predicts relevance from a post-SID relevance-token state via an MLP.
As shown in Table~\ref{tab:offline_corr}, UniGD consistently outperforms
both online alternatives across all metrics, demonstrating the advantage
of explicit query–ad representation matching. With a $60$M backbone
and only $1.1$M additional parameters, UniGD also performs comparably
to the offline teacher and achieves a marginally higher Pearson
correlation. These results show that UniGD preserves strong relevance estimation
within the generative retriever in production, avoiding a separately
deployed online relevance model.

\subsection{Ablation Study}
\label{sec:ablation}
\begin{figure}[t]
\centering
\includegraphics[width=\columnwidth]{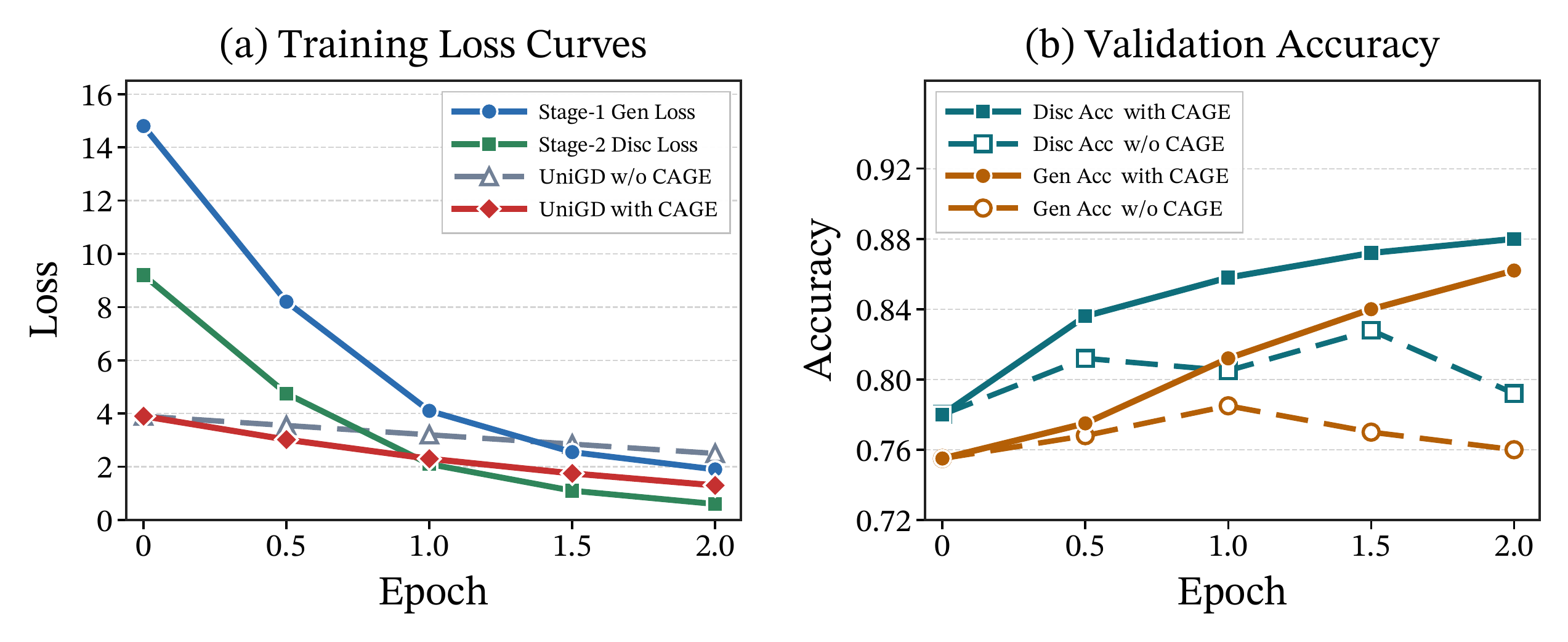}
\caption{Training dynamics of UniGD. (a) Training loss across the
three stages. (b) Generative and discriminative accuracy
during joint training with and without CAGE.}
\label{fig:online_training_dynamics}
\end{figure}

We ablate the components of UniGD on NQ320K and our industrial advertising
dataset. As shown in Table~\ref{tab:ablation}, we remove
CAM and replace CAGE with standard joint optimization on both datasets.
On our industrial dataset, we replace HAM with a shared
codebook, SID space, and generation head across all material types.
Removing CAM causes the largest degradation in Recall@1 and MRR@10,
demonstrating the importance of codebook-anchored representations and
explicit relevance estimation for candidate ranking. Removing CAGE
consistently degrades all metrics, while replacing HAM with shared
modeling reduces industrial performance, confirming the respective
benefits of conflict-aware optimization and material-specific modeling.
Figure~\ref{fig:online_training_dynamics}  examines the effect
of CAGE during joint optimization. With CAGE, both generative and
discriminative validation accuracy improve steadily; without it, they
decline after reaching intermediate peaks. These results suggest that CAGE mitigates optimization interference and stabilizes joint training.

\subsection{Industrial Online Evaluation}
\label{sec:industrial}

We conduct a one-week controlled online A/B test in the Kuaishou search
advertising system serving hundreds of millions of ad
candidates, where newly indexed ads account for $36.9\%$ of daily
traffic. All groups share the same serving environment and differ only
in their retrieval and relevance-estimation pipelines. We compare
UniGD with the production cascade of GR and an external relevance
model, GR + RelToken, and UniGD-Gen. UniGD-Gen retains UniGD's
generative component but replaces its internal discriminative scoring
with the production relevance model.

As shown in Table~\ref{tab:online_ab}, UniGD improves revenue and
new-ad impressions by $5.78\%$ and $24.60\%$, respectively, over the
production baseline, while reducing the Irrelevant Ratio. It also
reduces retrieval latency from $13.00$\,ms to $8.70$\,ms by replacing
the external relevance model with CAM lookup and lightweight internal
scoring. GR + RelToken substantially increases the Irrelevant Ratio to
$17.30\%$, indicating that a post-SID token state provides a less
reliable relevance signal than explicit query–ad representation
matching. UniGD's consistent advantages over GR + RelToken and
UniGD-Gen across all major metrics demonstrate that directly
integrating discriminative relevance estimation into generative
retrieval improves both effectiveness and serving efficiency.

\section{Conclusion}
We present UniGD, a unified generative-discriminative framework for
industrial search advertising. UniGD couples autoregressive candidate
generation with explicit relevance estimation in a shared model,
where CAM reconstructs ad representations from semantic codebooks,
HAM accommodates heterogeneous ad materials, and CAGE coordinates the
two objectives during joint optimization. Experiments on public
benchmarks and large-scale industrial data demonstrate improvements
in generative retrieval and relevance estimation, while online A/B
tests in Kuaishou search advertising further validate its effectiveness in production.
Beyond search advertising, UniGD can potentially be applied to
recommendation and general information retrieval by coupling
generative candidate retrieval with task-specific discriminative
objectives, such as relevance estimation, preference prediction, and
ranking. Evaluating UniGD in these broader domains is a promising
direction for future work.

% References and End of Paper
\bibliography{IEEEabrv,anthology,custom}

\begin{thebibliography}{38}
\providecommand{\natexlab}[1]{#1}

\bibitem[{Bajaj et~al.(2016)Bajaj, Campos, Craswell, Deng, Gao, Liu, Majumder,
  McNamara, Mitra, Nguyen et~al.}]{bajaj2016msmarco}
Bajaj, P.; Campos, D.; Craswell, N.; Deng, L.; Gao, J.; Liu, X.; Majumder, R.;
  McNamara, A.; Mitra, B.; Nguyen, T.; et~al. 2016.
\newblock {MS MARCO}: A Human Generated Machine Reading Comprehension Dataset.
\newblock arXiv:1611.09268.

\bibitem[{Bevilacqua et~al.(2022)Bevilacqua, Ottaviano, Lewis, Yih, Riedel, and
  Petroni}]{bevilacqua2022autoregressive}
Bevilacqua, M.; Ottaviano, G.; Lewis, P.; Yih, W.-t.; Riedel, S.; and Petroni,
  F. 2022.
\newblock Autoregressive Search Engines: Generating Substrings as Document
  Identifiers.
\newblock In \emph{Advances in Neural Information Processing Systems},
  volume~35, 31668--31683.

\bibitem[{Chen et~al.(2025)Chen, Guo, Wang, Liang, Lv, Ma, Xiao, Xue, Zhang,
  Yang et~al.}]{chen2025onesearch}
Chen, B.; Guo, X.; Wang, S.; Liang, Z.; Lv, Y.; Ma, Y.; Xiao, X.; Xue, B.;
  Zhang, X.; Yang, Y.; et~al. 2025.
\newblock {OneSearch}: A Preliminary Exploration of the Unified End-to-End
  Generative Framework for E-commerce Search.
\newblock arXiv:2509.03236.

\bibitem[{Cheng et~al.(2025)Cheng, Dou, Zhu, and Li}]{cheng2025ddid}
Cheng, J.; Dou, Z.; Zhu, Y.; and Li, X. 2025.
\newblock Descriptive and Discriminative Document Identifiers for Generative
  Retrieval.
\newblock In \emph{Proceedings of the AAAI Conference on Artificial
  Intelligence}, volume~39.

\bibitem[{Dai et~al.(2025)Dai, Tang, Wu, Wang, Zhu, Chen, Hong, Zhao, Fu, Wu
  et~al.}]{onepiece2024}
Dai, S.; Tang, J.; Wu, J.; Wang, K.; Zhu, Y.; Chen, B.; Hong, B.; Zhao, Y.; Fu,
  C.; Wu, K.; et~al. 2025.
\newblock {OnePiece}: Bringing Context Engineering and Reasoning to Industrial
  Cascade Ranking System.
\newblock arXiv:2509.18091.

\bibitem[{Deng et~al.(2025)Deng, Wang, Cai, Ren, Hu, Ding, Luo, and
  Zhou}]{deng2025onerec}
Deng, J.; Wang, S.; Cai, K.; Ren, L.; Hu, Q.; Ding, W.; Luo, Q.; and Zhou, G.
  2025.
\newblock {OneRec}: Unifying Retrieve and Rank with Generative Recommender and
  Preference Alignment.
\newblock arXiv:2502.18965.

\bibitem[{Fu et~al.(2026)Fu, Wu, Yuan, Cao, Huang, Yan, Zheng, Zhang, Zhou, Wu
  et~al.}]{rankgr}
Fu, K.; Wu, C.; Yuan, K.; Cao, B.; Huang, D.; Yan, Y.; Zheng, J.; Zhang, J.;
  Zhou, S.; Wu, J.; et~al. 2026.
\newblock {RankGR}: Rank-Enhanced Generative Retrieval with Listwise Direct
  Preference Optimization in Recommendation.
\newblock arXiv:2602.08575.

\bibitem[{Guo et~al.(2025)Guo, Chen, Wang, Yang, Lei, Ding, and Li}]{onesug}
Guo, X.; Chen, B.; Wang, S.; Yang, Y.; Lei, C.; Ding, Y.; and Li, H. 2025.
\newblock {OneSug}: The Unified End-to-End Generative Framework for E-commerce
  Query Suggestion.
\newblock arXiv:2506.06913.

\bibitem[{Karpukhin et~al.(2020)Karpukhin, O{\u{g}}uz, Min, Lewis, Wu, Edunov,
  Chen, and Yih}]{karpukhin2020dense}
Karpukhin, V.; O{\u{g}}uz, B.; Min, S.; Lewis, P.; Wu, L.; Edunov, S.; Chen,
  D.; and Yih, W.-t. 2020.
\newblock Dense Passage Retrieval for Open-Domain Question Answering.
\newblock In \emph{Proceedings of the 2020 Conference on Empirical Methods in
  Natural Language Processing}, 6769--6781.

\bibitem[{Kwiatkowski et~al.(2019)Kwiatkowski, Palomaki, Redfield, Collins,
  Parikh, Alberti, Epstein, Polosukhin, Devlin, Lee
  et~al.}]{kwiatkowski2019natural}
Kwiatkowski, T.; Palomaki, J.; Redfield, O.; Collins, M.; Parikh, A.; Alberti,
  C.; Epstein, D.; Polosukhin, I.; Devlin, J.; Lee, K.; et~al. 2019.
\newblock Natural Questions: A Benchmark for Question Answering Research.
\newblock \emph{Transactions of the Association for Computational Linguistics},
  7: 452--466.

\bibitem[{Li et~al.(2024{\natexlab{a}})Li, Wang, Chen, Nie, Qiu, Tang, Liu, and
  Zhuo}]{li2024genrpo}
Li, M.; Wang, H.; Chen, Z.; Nie, G.; Qiu, Y.; Tang, G.; Liu, L.; and Zhuo, J.
  2024{\natexlab{a}}.
\newblock Generative Retrieval with Preference Optimization for E-commerce
  Search.
\newblock arXiv:2407.19829.

\bibitem[{Li et~al.(2024{\natexlab{b}})Li, Yang, Wang, Wei, and
  Li}]{li2024ltrgr}
Li, Y.; Yang, N.; Wang, L.; Wei, F.; and Li, W. 2024{\natexlab{b}}.
\newblock Learning to Rank in Generative Retrieval.
\newblock In \emph{Proceedings of the AAAI Conference on Artificial
  Intelligence}.
\newblock ArXiv:2306.15222.

\bibitem[{Mekonnen, Tang, and de~Rijke(2025)}]{mekonnen2025ddro}
Mekonnen, K.~A.; Tang, Y.; and de~Rijke, M. 2025.
\newblock Lightweight and Direct Document Relevance Optimization for Generative
  Information Retrieval.
\newblock In \emph{Proceedings of the 48th International ACM SIGIR Conference
  on Research and Development in Information Retrieval (SIGIR)}.

\bibitem[{Ni et~al.(2022)Ni, Abrego, Constant, Ma, Hall, Cer, and
  Yang}]{ni2021sentence}
Ni, J.; Abrego, G.~H.; Constant, N.; Ma, J.; Hall, K.; Cer, D.; and Yang, Y.
  2022.
\newblock Sentence-{T5}: Scalable Sentence Encoders from Pre-trained
  Text-to-Text Models.
\newblock In \emph{Findings of the Association for Computational Linguistics:
  ACL 2022}, 1864--1874.

\bibitem[{Nogueira and Lin(2019)}]{nogueira2019doc2query}
Nogueira, R.; and Lin, J. 2019.
\newblock From doc2query to {docTTTTTquery}.
\newblock \emph{Online preprint}.

\bibitem[{Pang et~al.(2025)Pang, Yuan, He, Fang, Xie, Qu, Jiang, Peng, Lin, Law
  et~al.}]{pang2025gram}
Pang, M.; Yuan, C.; He, X.; Fang, Z.; Xie, D.; Qu, F.; Jiang, X.; Peng, C.;
  Lin, Z.; Law, C.; et~al. 2025.
\newblock Generative Retrieval and Alignment Model: A New Paradigm for
  E-commerce Retrieval.
\newblock In \emph{Companion Proceedings of the ACM Web Conference 2025},
  413--421.

\bibitem[{Qu et~al.(2021)Qu, Ding, Liu, Liu, Ren, Zhao, Dong, Wu, and
  Wang}]{rocketqa}
Qu, Y.; Ding, Y.; Liu, J.; Liu, K.; Ren, R.; Zhao, W.~X.; Dong, D.; Wu, H.; and
  Wang, H. 2021.
\newblock RocketQA: An Optimized Training Approach to Dense Passage Retrieval
  for Open-Domain Question Answering.
\newblock In \emph{Proceedings of NAACL-HLT}, 5835--5847.

\bibitem[{Rafailov et~al.(2023)Rafailov, Sharma, Mitchell, Ermon, Manning, and
  Finn}]{rafailov2023dpo}
Rafailov, R.; Sharma, A.; Mitchell, E.; Ermon, S.; Manning, C.~D.; and Finn, C.
  2023.
\newblock Direct Preference Optimization: Your Language Model is Secretly a
  Reward Model.
\newblock arXiv:2305.18290.

\bibitem[{Raffel et~al.(2020)Raffel, Shazeer, Roberts, Lee, Narang, Matena,
  Zhou, Li, and Liu}]{raffel2020t5}
Raffel, C.; Shazeer, N.; Roberts, A.; Lee, K.; Narang, S.; Matena, M.; Zhou,
  Y.; Li, W.; and Liu, P.~J. 2020.
\newblock Exploring the Limits of Transfer Learning with a Unified Text-to-Text
  Transformer.
\newblock \emph{Journal of Machine Learning Research}, 21(140): 1--67.

\bibitem[{Rajput et~al.(2023)Rajput, Mehta, Singh, Keshavan, Heldt, Hong, Tay,
  Tran, Samost, Kula et~al.}]{rajput2023recommender}
Rajput, S.; Mehta, N.; Singh, A.; Keshavan, R.~H.; Heldt, L.; Hong, L.; Tay,
  Y.; Tran, V.~Q.; Samost, J.; Kula, M.; et~al. 2023.
\newblock Recommender Systems with Generative Retrieval.
\newblock In \emph{Advances in Neural Information Processing Systems},
  volume~36.

\bibitem[{Robertson et~al.(1995)Robertson, Walker, Jones, Hancock-Beaulieu, and
  Gatford}]{robertson1995bm25}
Robertson, S.~E.; Walker, S.; Jones, S.; Hancock-Beaulieu, M.~M.; and Gatford,
  M. 1995.
\newblock Okapi at {TREC-3}.
\newblock In \emph{Proceedings of the Third Text REtrieval Conference
  (TREC-3)}, 109--126.

\bibitem[{Sun et~al.(2023)Sun, Yan, Chen, Wang, Zhu, Ren, Chen, Yin, de~Rijke,
  and Ren}]{sun2023gentoken}
Sun, W.; Yan, L.; Chen, Z.; Wang, S.; Zhu, H.; Ren, P.; Chen, Z.; Yin, D.;
  de~Rijke, M.; and Ren, Z. 2023.
\newblock Learning to Tokenize for Generative Retrieval.
\newblock In \emph{Advances in Neural Information Processing Systems
  (NeurIPS)}.

\bibitem[{Sun et~al.(2026)Sun, Huang, Guan, Luo, Tang, Gai, and Zhou}]{grank}
Sun, Y.; Huang, S.; Guan, Z.; Luo, Q.; Tang, R.; Gai, K.; and Zhou, G. 2026.
\newblock {GRank}: Towards Target-Aware and Streamlined Industrial Retrieval
  with a Generate-Rank Framework.
\newblock In \emph{Proceedings of the ACM Web Conference 2026}, 1--11.

\bibitem[{Tang et~al.(2024)Tang, Zhang, Guo, de~Rijke, Chen, and
  Cheng}]{tang2024gr2}
Tang, Y.; Zhang, R.; Guo, J.; de~Rijke, M.; Chen, W.; and Cheng, X. 2024.
\newblock Generative Retrieval Meets Multi-Graded Relevance.
\newblock In \emph{Advances in Neural Information Processing Systems},
  volume~37.

\bibitem[{Tay et~al.(2022)Tay, Tran, Dehghani, Ni, Bahri, Mehta, Qin, Hui,
  Zhao, Gupta et~al.}]{tay2022transformer}
Tay, Y.; Tran, V.; Dehghani, M.; Ni, J.; Bahri, D.; Mehta, H.; Qin, Z.; Hui,
  K.; Zhao, Z.; Gupta, J.; et~al. 2022.
\newblock Transformer Memory as a Differentiable Search Index.
\newblock In \emph{Advances in Neural Information Processing Systems},
  volume~35, 21831--21843.

\bibitem[{van~den Oord, Vinyals, and Kavukcuoglu(2017)}]{vqvae2017}
van~den Oord, A.; Vinyals, O.; and Kavukcuoglu, K. 2017.
\newblock Neural Discrete Representation Learning.
\newblock In \emph{Advances in Neural Information Processing Systems},
  volume~30.

\bibitem[{Wang et~al.(2022)Wang, Hou, Wang, Miao, Wu, Chen, Xia, Chi, Zhao, Liu
  et~al.}]{wang2022neural}
Wang, Y.; Hou, Y.; Wang, H.; Miao, Z.; Wu, S.; Chen, Q.; Xia, Y.; Chi, C.;
  Zhao, G.; Liu, Z.; et~al. 2022.
\newblock A Neural Corpus Indexer for Document Retrieval.
\newblock In \emph{Advances in Neural Information Processing Systems},
  volume~35, 25600--25614.

\bibitem[{Xiong et~al.(2021)Xiong, Xiong, Li, Tang, Liu, Bennett, Ahmed, and
  Overwijk}]{xiong2021ance}
Xiong, L.; Xiong, C.; Li, Y.; Tang, K.-F.; Liu, J.; Bennett, P.~N.; Ahmed, J.;
  and Overwijk, A. 2021.
\newblock Approximate Nearest Neighbor Negative Contrastive Learning for Dense
  Text Retrieval.
\newblock In \emph{International Conference on Learning Representations
  (ICLR)}.

\bibitem[{Xue et~al.(2026)Xue, Liu, Wang, Sun, Wang, Zhang, Shi, Xu, Sha, Liu
  et~al.}]{gr4ad2024}
Xue, B.; Liu, D.; Wang, L.; Sun, M.; Wang, P.; Zhang, P.; Shi, S.; Xu, T.; Sha,
  Y.; Liu, Z.; et~al. 2026.
\newblock Generative Recommendation for Large-Scale Advertising.
\newblock arXiv:2602.22732.

\bibitem[{Zeng et~al.(2024)Zeng, Luo, Jin, Sarwar, Wei, and
  Zamani}]{zeng2024ripor}
Zeng, H.; Luo, C.; Jin, B.; Sarwar, S.~M.; Wei, T.; and Zamani, H. 2024.
\newblock Scalable and Effective Generative Information Retrieval.
\newblock In \emph{Proceedings of the ACM Web Conference (WWW)}.

\bibitem[{Zeng, Luo, and Zamani(2024)}]{zeng2024pag}
Zeng, H.; Luo, C.; and Zamani, H. 2024.
\newblock Planning Ahead in Generative Retrieval: Guiding Autoregressive
  Generation through Simultaneous Decoding.
\newblock In \emph{Proceedings of the 47th International ACM SIGIR Conference
  on Research and Development in Information Retrieval}.

\bibitem[{Zhan et~al.(2020)Zhan, Mao, Liu, Zhang, and Ma}]{zhan2020repbert}
Zhan, J.; Mao, J.; Liu, Y.; Zhang, M.; and Ma, S. 2020.
\newblock {RepBERT}: Contextualized Text Embeddings for First-Stage Retrieval.
\newblock arXiv:2006.15498.

\bibitem[{Zhang et~al.(2025)Zhang, Song, Lee, Guo, Wang, Li, Guo, Liu, Lian,
  and Chen}]{unigrf}
Zhang, L.; Song, K.; Lee, Y.~Q.; Guo, W.; Wang, H.; Li, Y.; Guo, H.; Liu, Y.;
  Lian, D.; and Chen, E. 2025.
\newblock Killing Two Birds with One Stone: Unifying Retrieval and Ranking with
  a Single Generative Recommendation Model.
\newblock arXiv:2504.16454.

\bibitem[{Zheng et~al.(2025)Zheng, Wang, Yang, Fan, Zhang, Wang, and
  Wang}]{zheng2025egav2}
Zheng, Z.; Wang, Z.; Yang, F.; Fan, J.; Zhang, T.; Wang, Y.; and Wang, X. 2025.
\newblock {EGA-V2}: An End-to-end Generative Framework for Industrial
  Advertising.
\newblock arXiv:2505.17549.

\bibitem[{Zhou, Dou, and Wen(2023)}]{zhou2023genrrl}
Zhou, Y.; Dou, Z.; and Wen, J.-R. 2023.
\newblock Enhancing Generative Retrieval with Reinforcement Learning from
  Relevance Feedback.
\newblock In \emph{Proceedings of the 2023 Conference on Empirical Methods in
  Natural Language Processing (EMNLP)}, 12481--12490.

\bibitem[{Zhou et~al.(2024)Zhou, Yao, Dou, Tu, Wu, Chua, and
  Wen}]{zhou2024roger}
Zhou, Y.; Yao, J.; Dou, Z.; Tu, Y.; Wu, L.; Chua, T.-S.; and Wen, J.-R. 2024.
\newblock {ROGER}: Ranking-Oriented Generative Retrieval.
\newblock \emph{ACM Transactions on Information Systems}, 43(1).

\bibitem[{Zhou et~al.(2022)Zhou, Yao, Dou, Wu, Zhang, and Wen}]{zhou2022ultron}
Zhou, Y.; Yao, J.; Dou, Z.; Wu, L.; Zhang, P.; and Wen, J.-R. 2022.
\newblock Ultron: An Ultimate Retriever on Corpus with a Model-based Indexer.
\newblock arXiv:2208.09257.

\bibitem[{Zhuang et~al.(2022)Zhuang, Ren, Shou, Pei, Gong, Zuccon, and
  Jiang}]{zhuang2022dsiqg}
Zhuang, S.; Ren, H.; Shou, L.; Pei, J.; Gong, M.; Zuccon, G.; and Jiang, D.
  2022.
\newblock Bridging the Gap Between Indexing and Retrieval for Differentiable
  Search Index with Query Generation.
\newblock arXiv:2206.10128.

\end{thebibliography}

\end{document}